# IMAGINE: An Integrated Model of Artificial Intelligence-Mediated Communication Effects

Frederic Guerrero-Solé

Universitat Pompeu Fabra

Artificial Intelligence (AI) is transforming all fields of knowledge and production. From surgery, autonomous driving, to image and video creation, AI seems to make possible hitherto unimaginable processes of automation and efficient creation. Media and communication are not an exception, and we are currently witnessing the dawn of powerful AI tools capable of creating artistic images from simple keywords, or to capture emotions from facial expression. These examples may be only the beginning of what can be in the future the engines for automatic AI real time creation of media content linked to the emotional and behavioural responses of individuals. Although it may seem we are still far from there, it is already the moment to adapt our theories about media to the hypothetical scenario in which content production can be done without human intervention, and governed by the controlled any reactions of the individual to the exposure to media content. Following that, I propose the definition of the Integrated Model of Artificial Intelligence-Mediated Communication Effects (IMAGINE), and its consequences on the way we understand media evolution (Scolari, 2012) and we think about media effects (Potter, 2010). The conceptual framework proposed is aimed to help scholars theorizing and doing research in a scenario of continuous real-time connection between AI measurement of people's responses to media, and the AI creation of content, with the objective of optimizing and maximizing the processes of influence. Parasocial interaction and real-time beautification are used as examples to model the functioning of the IMAGINE process.

Keywords: artificial intelligence, media effects, real-time content generation, computer vision, communication theory, parasocial interaction, beautification, facial expression

The development of Ai technologies based on neural networks and deep learning is having a huge impact of fields of knowledge as diverse as neuroscience (Hassabis, Kumaran, Summerfield & Botvinick, 2017), genetics (Libbrecht & Noble, 2015), medicine (Topol, 2019), clinical decision support (Shortliffe & Sepúlveda, 2018), healthcare (Jiang, et al., 2017), drug design (Schneider et al., 2020) and chemistry synthesis (Struble et al., 2020), robotics (Laird, Lebiere, & Rosenbloom, 2017), decision-making for high-impact weather (McGovern, 2017), precision agriculture (Patrício & Rieder, 2018), achieving the sustainable development goals (Vinuesa, et al., 2020), or marketing (Davenport, Guha, Grewal & Bressgott, 2020), among other fields of knowledge and practice. The scope of applications of AI is vast, and it affects almost any human activity, regardless of the type of actions in which it is embedded. Ai is not only becoming a key agent in critical decisions on people's health or weather forecasting; it is also having a huge impact on fields that have been considered central characteristics of human intelligence, such as creativity (Boden, 1998), mediated communication (Kaplan & Haenlein, 2019), and artistic and media content creation.

The generation of creative ideas by AI is not new (Boden, 1998), but recent improvements of algorithms and techniques, such as transformer-based pretrained text-to-image synthesis models, and the availability of big datasets have moved AI creativity a step forward. The future impact of these new technologies on activities like art, fashion, videogames, or media content production is still unknown; however, recent developments point towards an acute transformation of creative activities and media industries. Drawing on the most recent studies on AI and communication (Sundar & Lee, 2022; Dehnert & Mongeau, 2022; Endacott & Leonardi, 2022), the objective of this paper is to theorize about the impact of AI on the study of media effects. I propose an integrated model of artificial intelligence-mediated communication effects, in which an AI agent (AA-receptor) monitors the process of reception of the stimuli, a second integrated AI agent (AA-creator) control the production in real-time, based on the inputs of the first agent, and a third AI agent (AA-negotiator) deals with the goals that are considered to be achieved on the receiver's mind. I discuss the philosophical and cultural roots of this model, and the theoretical and methodological consequences of the embedment of AI agents in the process of mediated communication. I finally relate the model to classical theorizations of the impact of media on individuals, and propose it as a starting point for a new theorization on media effects, in which artificial agents play a fundamental role.

## Artificial Intelligence. A definition

Artificial Intelligence (AI) is nowadays one of the most promising fields in science. By means of rigorous mathematical theories (Russell & Norvig, 2022), AI has showed its capabilities in solving complex problems such as autonomous driving or chemical synthesis, and has turned into a critical tool for the future of a wide variety of industries and organizations. AI is defined as a technology of computation capable of interpreting external data, learn from them, use the resulting learning to decide the optimal actions to achieve specific goals (HLEG AI, 2019) and to flexibly adapt to the changes in the environment (Haenlein & Kaplan, 2019). AI can learn from experience, recognize patterns (Endacott & Leonardi, 2022), and make decisions as if it was human, without human intervention. This learning process has been boosted in the last years by the access to large datasets (Duan Edwards & Dwivedi, 2019).

AI allows the automation, iteration and optimization of processes that, in the past, were only done by humans (Eiband & Buschek, 2020). In this sense, AI is to be considered as a computational rational agent that acts given inputs or percepts to achieve the best and optimal expected outcome (Russell & Norvig, 2022).

The philosophical understanding of the human mind as a machine and the laws of mathematics have defined the evolution of artificial intelligence from the 1950s. AI is, consequently, the consummation, the culmination of rational thinking and logical positivism, that understand that any kind of knowledge can be characterized by logical laws connected to observation by means of sensors (Russell & Norvig, 2022). Accordingly, we define an intelligent agent (IA) as a technology of sensors or percepts that can measure variables in a particular environment or object.

AI is commonly divided in two types: narrow AI, that refers to intelligent technologies capable of performing specific tasks, and general AI, capable of reason, plan, solve problems and, finally, mimic human intelligence. In the particular case of communication, narrow AI has proven to be successful in performing tasks such as optimal searching, voice assistance, speech recognition, spam fighting, machine translation (Russell & Norvig, 2022), or text-to-image tools. The impulse of natural language processing and AI-assisted creation has been facilitated by mathematical developments such as neural (Lecun, Bottou, Bengio & Haffner, 1998), transformer and generative adversarial networks, deep learning (LeCun, Bengio & Hinton, 2015; Wu, Gong, Ke, Liang, Li, Xu, Liu & Zhong, 2022), or diffusion models (Sohl-Dickstein, Weiss, Maheswaranathan & Ganguli, 2015). In the particular case of content creation and text-to-image tools, diffusion models have proved to be particularly successful in creating artistic outputs from simple strings of natural language, and AI-systems such as DALL-E, Stable Diffusion or MidJourney are becoming of common use for illustration and digital design.

However, AI can be introduced not only in the process of creation, but in the whole process of mediated communication. Hancock, Naaman and Levy (2020) define AI mediated communication (AIMC) as the communicative process in which an artificial agent modifies, augments or generates message to achieve certain goals. According to the authors, the dimensions of Artificial Intelligence Mediated Communication (AIMC), are magnitude, media type, optimization goal, autonomy, and role orientation. Magnitude refers to the changes enacted by the Artificial Intelligence actor, the autonomy to the degree of freedom to operate without the sender's intervention, and role orientation refers to the agency, on the sender or the receiver. But this definition is exclusively focused on the process of creation of the message, on the creative tasks of AI, without considering how AI can also impact the process of reception and, more importantly, in the process of monitoring, processing, analyzing and reusing the data from people's emotional and cognitive responses to their exposure to AI-assisted or generated media content. Consequently, optimization goal, as it is understood by Hancock, Naaman and Levy (2020) must consider not only the message, but also the impact of the message on the receiver, which is the main focus of media effects.

The impact of AI on mediated communication

In summer, 2022, Human Communication Research published a special issue in which Sundar and Lee (2022) urged about the necessity to rethink communication and adapt the discipline to the era of AI. The special issue contained researches related to the intersection between AI and persuasion (Dehnert & Mongeau, 2022), the interaction between humans and AI social chatbots (Brandtzaeg, Skjuve & Følstad, 2022), the consequences of delegating self-presentation and impression management to AI agents (Endacott & Leonardi, 2022), the role played by AI agents as fact-checkers (Banas, et al., 2022), or the role played by AI bots in the spreading of information in social networks (Duan et al., 2022). Although previous works had already delved into the interaction between AI and communication (Guzman, 2019; Donath, 2021), the special issue was particularly helpful in stressing the different actions of the communicative process in which AI could or can play a significant role.

Dehnert & Mongeau (2022) propose a theoretical approach on the contribution of AI in the process of persuasion, proposing a definition of what is to be considered AI-based persuasion: a symbolic process in which AI generates, augments, or modifies a message that is designed to provoke an effect on humans. As stated by Donath (2021), although AI systems using data for persuasion are still early in this development, deep and machine learning technologies may soon be able to create persuasive content. Brandtzaeg, Skjuve, and Følstad (2022) analysed the friendship-like relationship that people establish with machines, finding that those interactions stimulate experiences that are similar to those of face-to-face communication. Endacott and Leonardi (2022) studied the consequences of the cession of control of impression management to AI agents that make decisions on behalf of their principals. Banas, et al. (2022) found that human (crowdsourced) and AI fact-checking agents' agreements have a strong effect on people's credibility judgements (consensus heuristic), regardless of the judgment of both agents. As a norm, people privilege those agents that are in tune with the heuristic (human/AI) they endorse, only when the agent indicates that the information is true. Finally, Duan et al. (2022) show that AI selectively amplifies and emphasizes certain topics in social networks debates, and responds to the content created by human and news media outlets. The authors warn that in the future, when bots interactions will be indistinguishable from those of human agents, AI can systematically introduce biases and misinformation.

All these contributions lead to Sundar and Lee (2022) to propose a classification of the involvement of AI in the process of Human Communication. They define the four main functions played by AI: creator, converser, curator and co-author (see Table 1).

**Table 1.** Classification of AI's Involvement in Human Communication, according to Sundar and Lee (2022).

|              | Mass (One-to-Many)                                          | Interpersonal (One-to-One)                        |
| ------------ | ----------------------------------------------------------- | ------------------------------------------------- |
| Communicator | AI Creator                                                  | AI Converser                                      |
|              | (e.g., Virtual Influencer, AI Anchor, Robot Reporter)       | (e.g., Chatbot, Virtual Assistant, Smart Speaker) |
| Mediator     | AI Curator                                                  | AI Co-Author                                      |

| (e.g., Content Moderator, AI Fact-checker, Recommendation System) | (e.g., Auto-Completion, Auto-Correction, Automated Response Suggestion) |

According to Sundar and Lee (2022), AI performs tasks of creating one-to-one or one-to-may messages, or of mediating in the communication process, by assisting humans or by curating contents and assisting humans in the process of message creation. However, the authors do not consider the role that AI technologies can play in the process of assessing the effects of AI-based communication on humans.

## The role of AI in the analysis of media contents and communication effects

Research on the impact of media contents on people has had a huge impact on the discipline of communication since the 1940s (Lazarsfeld, Berelson & Gaudet, 1948). Mass media effects deals with the changes in outcomes due to the exposure to media contents (Potter, 2011; Potter, 2017), being these outcomes knowledge, beliefs, attitudes, affects, physiological responses and behaviours (Potter, 2011). For Berger and Chaffee (1988), all mass communication studies more or less explicitly try to explain the effects of the media in one form or another.

From the first researches onwards, we have witnessed an accumulative theoretical progress on the effects of media on people (Neuman & Guggenheim, 2011). However, one of the main complexities for media effects research has traditionally been the methods for reliably measure the changes in the outcomes. Research includes different methodologies such as ethnography, observation, visual orientation (Lorch, 1994), focus group, interview, content analysis, survey, experiments and statistical analysis (Bryant & Zillmann, 1984; Hansen & Machin, 2013; Berger, 2000; Tewksbury & Scheufele, 2009), biological measures (Marczyk, DeMatteo & Festinger, 2005) or psychophysiological interaction analysis (Lang, Potter & Bolls, 2009; Kubitschko & Kaun, 2016), among others. In general, literature has been dominated by participants' self-reports, that Potter (2011) considers the most appropriate method to assess people's changes due to their exposure to media contents. But self-reporting often leads to the introduction of several biases, based on the fact that people often report different from actual behaviours, and the obsession with statistical significance (Potter, 2011).

To overcome the biases and the obstacles that stem from the use of methods that do not guarantee the truthfulness of the data collected, researchers have been applying new methods and technologies to measure people's responses and changes to their exposure to media contents. In this sense, Potter and Bolls (2012) pointed out that future of research in communication is to shift from a focus on creation to a focus on understanding the way human beings process media messages (Potter & Bolls, 2012) by means of complex methodologies that may include the impact on the way people react to the exposure to media due to the changes in the social environment. Probably, this is still an underdeveloped area, but there have been promising improvements in the field of computer vision, neuromarketing (Ariely & Berns, 2010) and other methodologies whose aim in to capture the effects of media contents on people's beliefs, attitudes, affects or behaviours.

In spite of being still a lot of room for the improvement of these tools, the study of people's responses to assess the effects on individuals has been one of the driving forces for scholars in communication. At the same time, the findings have been used to improve strategies to reach certain objectives, such as the increase of attention or the changes in attitudes and behaviours. These new and sophisticated methods include electrodermal response (EDR) and electroencephalograph (EEG), that provide of time-sequenced data to understand the dynamic process of communication (Watt, 1994), or computer vision and hearing techniques to recognize human's emotions (Picard, 1997).

But in a communication environment in which people increasingly interacting with human-like artificial agents (Guzman, 2019), AI can also play a role in the process of measuring people's responses to content generated artificially. Donath (2021) considers that AI systems are improving faster and faster, and that they know more and more about human complexity and emotions. In particular, if we consider that techniques such as EEG are not fully effective in measuring people's responses to media stimuli because of the noise (Pickering, 2010). Recent developments in functional magnetic resonance imaging (fMRI) (Turner, Huskey & Weber, 2019) or in brain recording suggest that in the future it will be possible to continuously decode language from non-invasive methods (Tang, LeBel, Jain & Huth, 2022). Leading companies in the sector of digital communication, such as Meta (Facebook, 2022) are already developing non-invasive tools to record brain activity and decode language, based on electroencephalography (EEG) and magnetoencephalography (MEG), using a model that consisted of a standard deep convolutional network (Défossez, Caucheteux, Rapin, Kabeli & King, 2022).

It is obvious that there is still a long way to go in the development of technologies and methods that can allow researchers to measure real and truthful cognitive and emotional states of people, if this is possible (Crawford, 2021; Heaven, 2020). In this sense, Crawford (2021) points out the controversy in relation to the automatic detection of facial expressions. However, the ontological and epistemological doubts about what an emotion is, or what is the relationship between facial expressions and emotions do not stop the industry from its objectives: to finally control the human brain by means of observation and stimulation.

In sum, nowadays brain-computer and computer vision technologies for monitoring people's responses to media messages allow a continuous measure and analysis of real-time data and a real-time control (Pickering, 2010) of the communicative process. Consequently, we can define the integrated model of AI mediated communication as the assembly of AI agents for content creation with AI agents for responses measuring and analysing, giving rise to a closed cycle of communication in which people's outputs become inputs for content generation.

## IMAGINE: An integrated model of AI mediated communication

For Guzman & Lewis (2020), the present context of narrowing distances between human communication and AI is due to the fact that now AI is becoming an agent that interacts with human beings. According to the authors, these interactions do not fit the classical theoretical frameworks within the discipline of communication. As it has been mentioned, mathematical progress and the availability of large datasets have boosted the development of AI tools for creating digital contents in real-time. Neural networks, transformer and generative adversarial networks, diffusion models and deep learning (Lecun, Bottou, Bengio & Haffner, 1998; LeCun, Bengio & Hinton, 2015; Wu, Gong, Ke, Liang, Li, Xu, Liu & Zhong, 2022; Sohl-Dickstein, Weiss, Maheswaranathan & Ganguli, 2015) have become the basis for the creation of text-to-image or text-to-video tools, that can be used nowadays by almost anyone. These tools commonly use text (prompts) as the inputs for creating the content. However, in a near future any input could be used to tell the AI create any kind of digital content, including texts, images, videos and sound.

All communication is a continuous process of interaction between the message and the receiver (Lang, 2009, 2000). Lang proposes the dynamic processes perspective, in which mental processes and psychological states evolve through the interaction with media contents. In particular, data provided by the AI tools for measuring and analysing people's response to media messages can be a source or input for creating new content. This is the main idea underlying the proposed integrated model of AI mediated communication: AI can play the role of creator and curator, but also the role of control of the reaction to the creation and the curation (Table 2).

**Table 2.** Additional AI's Involvement in Human Communication

| | |
|---|---|
| Receiver | AA Receptor |
| | (e.g., fMRI, EMG) |
| Goal Negotiator | AA Negotiator |
| | (e.g., algorithms) |

## The characteristics of IMAGINE

The integration of artificial agents (AA) radically changes the nature and characteristics of the communicative process. The Integrated model of AI mediated communication (IMAGINE) proposes a model with three main types of AA: an AA with the task of creating media contents (output) in real-time by means of a certain input (AA-creator); an AA with the task of measuring (output) the emotional and cognitive states of the individual exposed to a media content (input) (AA-receptor); and, finally, an AA with the task of negotiating between the AA-creator and the AA-receptor, considering the goals of the communicative process.

**Figure 1.** Proposed architecture for the Integrated model of AI mediated communication.

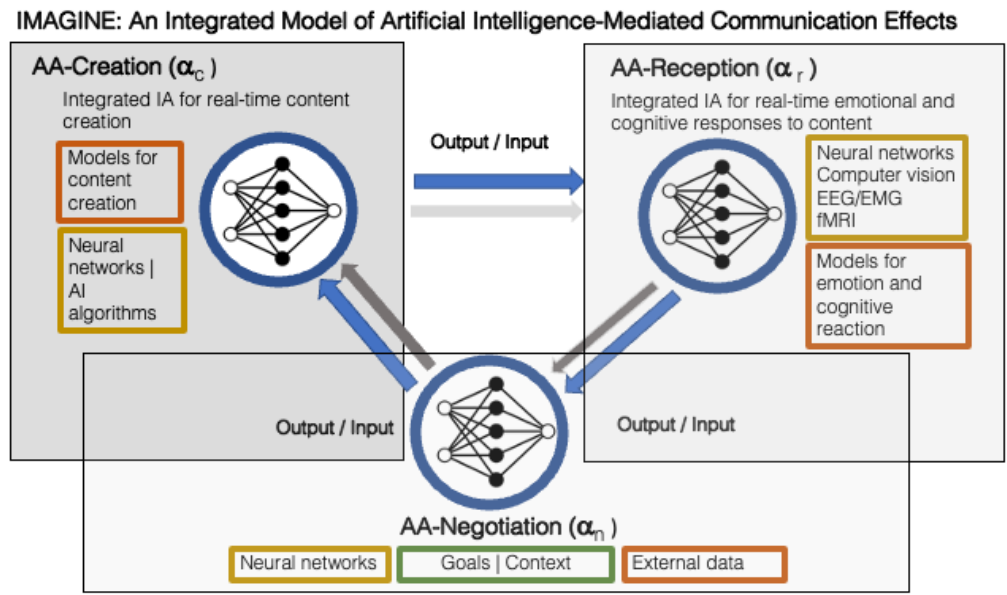

The characteristics of the whole IMAGINE model can be summarized in the following way:

## Model-based, data-centred and algorithmic control

The integrated model is data and algorithmic-centred. The creation of content is based on machine learning models and artificial neural networks, as expression of technologies that allow the creation of almost any visual and sound content from inputs and models. The content itself doesn't exist, but it is a creation of the algorithms, based on the data in models and any other information provided by the AA-negotiator. The AA-receptor has a similar structure, and it is based on the information provided by the integrated methods for measuring people's cognition and emotion (being MEG, EEG, fMRI and computer vision) and models, that transform the measure of variables into something understandable by the AA-negotiator. Consequently, the model is a model based on a cycle of data transformation.

## Integration and bidirectionality

In this sense, the main characteristic of the IMAGINE is the intimate integration of the processes of creation and reception, a fact that has been largely stressed by scholars of different traditions in the study of communication. This integration means that not only emission has an impact on reception, but also reception on emission, and in an explicit way. Obviously, it doesn't change classical questions of communication studies in relation to long-term effects, such as selective interpretation or retention, but in the short-term, reception and emission are intertwined and can be even interchangeable.

## Real-time and synchronicity

As noted by Pickering (2010) nowadays devices allow the control of the communicative process in real-time. It means that the IMAGINE is a model in which creation and reception are decided and measured respectively in real-time. Recent researches shown how through intracortical BCI it is possible, for instance, to decode from neural activity attempts to handwrite and transforms them into text in real-time (Willett, Avansino, Hochberg, Henderson & Shenoy, 2021). The decisions for the creation of content come from the continuous measure of the reception of the previous content and the potential expectations of the individual, that are provided by the AA-receptor. It leads to a synchronous assembly of both processes: the output of one of the processes becomes the input of the other process in real-time.

## Goal-orientation and optimization

As any AI tool, the IMAGINE is goal-oriented. The goals are the main components of the negotiation between the AI creation and the AI reception agents in the so-called AI Link (see Figure 1 These goals can be related to increasing attention, changing beliefs, attitudes and behaviours, or to help people with certain cognitive diseases or neurological neuromuscular disorders (Nicolas-Alonso & Gómez-Gil, 2012) as it is the objective of the Facebook (2022) research nowadays, in their effort for decoding speech from brain activity for people with traumatic brain injuries.

By means of AI tools, the whole process of communication is oriented to the optimization of the impact of the media content on the receptor (Dehnert & Mongeau, 2022).

Another aspect to consider is the nature of the goals of the model. The goals can be set by the individual herself, or by a third-party. The goals of the third-party can be, in a very general and non-exhaustive way, related to persuasion (marketing purposes, that can include effects on the beliefs, attitudes and behaviours) or cognition (for health purposes). These goals require the control of the variables measured by the AA-receptor, and the consequent establishment of certain thresholds or desired values of the variables in time. When the exposure to AI-generated contents leads to the achievement of the goals, then the process is convergent (Kincaid, 1980). If goals are not achieved, then the process is said to be divergent.

On the other hand, if the goals are set by the individual, IMAGINE may be related to content creation (it is not the content that controls the brain, but the brain that controls de content, as in the case of BCI devices for talking or writing). They could also be related to precise and explicit uses and gratifications obtained from the exposure to the AI-generated contents. The same concepts of convergence and divergence may be applied here.

## Interaction and negotiation

In the IMAGINE model, the interaction is not only of the receptor with the creator. The process of negotiation between the outputs and the inputs of the two agents (creation and reception) carried on by the AA-negotiator can be also understood as interactive. These negotiation orients the creation of the new content that the individual is going to be exposed to.

In this sense, IMAGINE goes beyond the most classical approach to interaction and postulates a two-way interaction in which the creator influences the receptor and vice versa.

The AA-negotiator knows about the relationship between the exposure to the content created and the effects that it may have on the receptor, that are measured by the AA-receptor.

## Adaptation and continuity

AI agents have been defined as the creators, modifiers of the message, on behalf of the sender, or as agents for the analysis of certain data related to the receiver, such as communication history or personal information; however, they have not been explicitly defined as the analysers and controllers of the whole process of communication and, in particular, of the outcomes on the receiver, i.e., the values of the variables related to their changes in beliefs, attitudes and behaviours, and any other outcome related to media effects. Dehnert and Mongeau (2022) defined persuasive AI as the development of a message, or a part of it, by an artificial agent, with a goal of influencing a response. Our approach goes beyond, since the AI agent also controls the response, and can adapt to the discrepancy between the measured values and the intended ones. Thus, the definition of the **integrated model** is in tune with the definition of intelligence: automated content, interaction with the individual through the gathering and processing of the data, and adaptation of the generation of new content according to the goals of the exposure to AI-generated content.

Consequently, within the IMAGINE model, the human brain can be understood as an environment in which an adaptative mechanism (AI) has to survive by achieving certain objectives. But, at the same time, the whole process of creation can be understood as subordinated to the outputs of the AA-receptor that, after negotiation, finally become inputs for the AA-creator. In that sense, the IMAGINE can be a model focused on the caption and the achievement of certain goals of media effects, or a tool for creating media contents that lately can be stimuli for other subjects. It is, in nature, an interactive process between the intelligence for the creation of contents and the intelligence from the reception, that can be influenced by the own thoughts of the receptor.

## Uniqueness and ephemerality

Being the model an architecture in which content is created in real-time, and depends on the scores of the variables measured by the AA-receptor, this content is, by nature, unique and ephemeral and leaves no traces. Although the conditions may be similar to those of the social networks, there is a relevant difference in the fact that they are goal-oriented and the creation is performed by the model itself, and not by a network of users that share their content.

## Brain-centred and non-invasion

AA-receptor uses algorithms for decoding neuron signals and transform them into an output that can be used as an input by the AA-negotiator to create a new output that becomes the input for the AA-creator. All this process is absolutely transparent to the user, by means of AI brain-computer interfaces (Wolpaw, Birbaumer, McFarland, Pfurtscheller & Vaughan, 2002; Nicolas-Alonso & Gómez-Gil, 2012) and computer vision for affective computing (Thies, Zollhofer, Stamminger, Theobalt & Niessner, 2016). Non-invasive devices allow the communication process to be performed in a natural way. Besides this, and in tune with the IMAGINE model, devices for the exposure to media contents or to immersive reality.

## AI and self-effects

The model pushes us to think about the effects of the algorithms, neural/convolutional networks and AI intelligences effects on people. This is probably one of the critical questions for mass media communication nowadays, since the characteristics of the IMAGINE model substantially differ from the hitherto classical approach to media effects. As it is already happening in social networks or search engines, any change in the algorithms can lead to different outcomes or effects, that can be used for commercial or political purposes. The SEME (authors, 2015) can be turned into a AIME, Artificial Intelligence Manipulation Effects, that can make research on effects even more complex and complicated as it is nowadays. Regulation and ethics will also be critical in the introduction of AI in the creation and reception of content.

A final question deals with the self-effects, the changes in beliefs, attitudes or behaviours provoked by the content created by oneself. In particular when the IMAGINE model works as a model for the creation of media contents.

## Accounting for the intelligence of AAs and models of communication

The proposed architecture of the integrated model of AI-mediated communication can lead us to the definition of three parameters that account for the level of intelligent of the AA defined. We will call these parameters $\alpha_c$ for creation, $\alpha_r$ for reception, and $\alpha_n$ for negotiation.

**Table 3.** Models of mediated communication depending on the level of intelligence applied to the three artificial agents of IMAGINE, creation, reception and negotiation (0 means the value of $\alpha$ tends to 0, 1 means the value of $\alpha$ tends to 1).

| $\alpha_c$ | $\alpha_r$ | $\alpha_n$ | Model of mediated communication |
|---|---|---|---|
| 0 | 0 | 0 | Broadcasting media model |
| 0 | 0 | 1 | Data-algorithmic model / Digital Platforms |
| 0 | 1 | 0 | Experimental model / fMRI, EEG |
| 1 | 0 | 0 | AI-creation model / Stable diffusion |
| 0 | 1 | 1 | AI-reception algorithmic experimental model |
| 1 | 0 | 1 | Algorithmic AI-creation model / Voice-to-image |
| 1 | 1 | 0 | No-effects oriented AI creation/experimental model |
| 1 | 1 | 1 | Ideal cycle of AI-mediated communication |

## Discussion

The embedment of AI within the process of mediated communication is transforming the way we understand the roles of creator and receptor, as well as the methods we use to measure media effects or the strategies for achieving these effects. Since media companies are interested in promising that they are vehicles for convincing people to purchase (Perse & Lambe, 2017), AI is being more and more used to define strategies of persuasion (Donath, 2021; Dehnert & Mongeau, 2022). Thus, in the field of communication, AI has been primarily understood as a way to support the creation and curation of content at a one-to-one or one-to-mass levels (Sundar & Lee, 2022). The IMAGINE model proposes the introduction of AI not only in the process of creation, but also in the reception and in the negotiation between creation and reception. It is a model in which different methods and technologies converge: from EEG or fMRI as non-invasive ways of reading people's responses to media contents, to the new developments in text-to-image, image-to-image, text-to-video or text-to-sound that are radically changing the same notion of creation. Machine learning, training models and neural and convolutional networks, among many other techniques, are facilitating both processes. The model also includes one of the classical characteristics of AI technologies, which is goal orientation. The goals can lead to the achievement of certain values of the variables measured on the subject's brain by the AA-receptor, or to create media content by the AA-creator as an objective defined by the subject.

Several theoretical implications can be drawn from the model. First and foremost, the IMAGINE model is linked to a tradition of thinking about the mediated communicative process as a driver to persuasion and to other media effects on people's beliefs, attitudes, emotions or behaviours. As it is defined, it can have a wide range of influences on people, depending on the goals specified in the AA-negotiator, and on the level of intelligence of both the AA-creator and the AA-receptor. In this sense, we can define an efficiency measure for the model, depending on its capacity to reach the goals. At one extreme, if we consider the model working as an ideal cycle (efficiency equal to 1), the IMAGINE can reproduce the schema of the hypodermic needle approach, with the AI-generated content having a powerful impact on the individual's brain, in form of changing of the variables that are in play in the negotiation between the AA- creator and the AA-receptor. On the other extreme, the model can reproduce the ineffective process of communication in which the individual has no motivation or interest, and that lead to no substantial changes in any of the variables measured. Linked to this approach, the IMAGINE model also draws the path of technological development in the field of communication, that is led by a desire to fully control the process of reception of the contents in media to maximize and optimize the final outcomes of people's exposure.

Another key consequence of the model is the potentially radical transformation of the whole process of measuring the effects of media on people. Since contents can be created and transformed in real-time in a continuous way, the analysis of discrete variables and the use of certain statistical tests based on these variables is, at the very least, useless. Instead of looking for differences and relationships between variables, which has been the basis of behaviourism and the media psychology approach (Potter & Bolls, 2012), research must be focused on the whole dynamics of changes on the individual's brain due to the continuous changes in the contents created by the AA-creator. In this sense, the AA-creator is defined as an intelligent agent that is capable of subtlety modify certain characteristics of the message in order to achieve or to approach the goal. Training and learning techniques will be key in the future to predict the transformations of the artificially generated contents to fit the objectives in the AA-negotiator.

In summary, AI is nowadays ruling the algorithms for distributing the contents generated by users in digital platforms and, at the same time, is allowing these users to create new images and text from scratch, and to modify in an almost infinite ways the contents they publish online. Consequently, Ai is introducing its power in the different elements that constitute the communication process. Bots become senders and co-creators, algorithms are defining the characteristics of the channel and having a huge impact on people's exposure to media content, and digital platforms are becoming spaces in which interactions are defined by AI tools.

The use of AI in communication may rise concerns about deception and manipulation (Hancock, Naaman & Levy, 2020). Being persuasion the main objective of media (Donath, 2021), to have an impact on people's beliefs, attitudes and behaviours, the intersection of AI and marketing suggest future threats for manipulation and deception of the audiences (Donath 2018, 2020). The question related to the goals of the exposure to contents must be also looked through the lens of ethics. Since any exposure to contents can be nowadays understood as an experiment, we have to be aware of the consequences for the individual of being subject to an intimate scrutiny of her thoughts and emotions. Companies such as Facebook (2022), developing instruments for reading people's thoughts, refer to traumas and diseases that incapacitate people for communication as the main reason of their technological progresses. However, it is obvious that these tools can be used for many different purposes that those of providing a way for expressing to people with brain damages. The ethics of AI-mediated communication and algorithmic fairness is and will be a necessary task to prevent abuse and misuse of a technology that, as a general norm, will created highly personalized and ephemeral contents with an unknown impact on people's cognition.

The definition also raises an important question about the use of AI in the process of communication. Whom is AI at the service of? As boyd and Crawford (2012) discussed the impact of big data *a*, I discuss the function of the embedment of AI technologies in digital platforms. For doing so, I contrapose the two classical approaches of mass communication research, i.e., media effects and uses and gratifications. The orientation of both approaches leads to the question about the objectives of interaction and communication. The creators of media messages define their own objectives, such as the increase of attention, or the change of beliefs or attitudes that may lead to certain behaviours of the receivers. These objectives can be assisted by AI-technologies, as it is the case of video-sharing digital platforms. On the other way around, from the users and gratifications approach, it is the individual the one that decides what are the objectives of her exposure to media contents. Individuals may expose to media to socialize, to pass time, to learn, or for any other objective-oriented reason. However, we may ask about the AI tools at her disposal. Conceiving the individual as rational, autonomous and active puts an obstacle to the use of AI technologies when exposing to media contents. People are intelligent enough not to need any additional support, and her needs are satisfied by the technology provided by the platforms. Within this frame, the gap between the intelligence at disposal of the creator and the receiver increases: AI is at the service of the media industry, pretending to be at the service of the user?

As a final conclusion, the IMAGINE model can help understanding the main drivers in the evolution of communication technology (Scolari, 2012). The history of media is also in the evolution of the methods for collecting data from media exposure. From careful and systematic observation to interviews and surveys, the collection of data in the past where segregated from the process of exposure itself. The media industry has been introducing the techniques of measuring within the technologies themselves, making them pervasive and, at the same time, transparent. It is frequently understood as a way of social expression, by liking or forwarding contents, or by following certain users.

## The ideal cycle of communication

The ideal cycle defines the conditions of a communication process, from the sender and the receiver perspectives. It links with the social physics approach that is intended to make understand social dynamics in terms of physical laws. This approach from physics has had a huge impulse with the emergence of network analysis in the field of communication. But it has its roots on the conception of society as a kind of matter formed by particles that statistically obey certain fixed laws that can be expressed by mathematical formulae. The virtuous cycle is, within this way of thinking, the horizon of the media industry that can be expressed in the following way: in ideal conditions, the pretended effects of media messages on people tend to their maximum values, while the unintended effects tend to vanish.

Media industry's ideal cycle is nowadays closer than ever. In particular, if we consider the present evolution of media technologies, in which the individual is proposed to be immerse in the digital platforms that reproduce the logics of everyday life. VR devices and environments such as the Metaverse are defining the future of people's exposure to contents. Being it accepted and adopted by people or not doesn't change the nature of one of the main objectives of the technology, by placing the individual in an environment that modifies her beliefs, attitudes or behaviours. And particularly behaviours and actions made in the environment itself, such as buying or voting, that have a social, economic meaning.

The ideal process of communication clearly refers to a perfect persuasion or manipulation machine, capable of influencing people in a stimulus-response way. Although the model is a theoretical one, and it lacks of an incontrovertible scientific and technical support, it may be considered as useful in illustrating the final point of logical positivism and rationality. The ideal cycle is not only a way to understand how AI can permeate the process of mediated communication. It is also the way to uncover the cultural and scientific roots of the infamous and mythical hypodermic needle model. I argue, thus, that the idea of the hypodermic needle model was not only a presumable wrong approach of the powerful impact of media messages on people, based or influenced by behaviourism, but a (or the) driving force of media's technological innovation. Consequently, we should doubt about the discourses that attributed the popularisation of the paradigm of the powerful effects to the audience's and the theorists' ignorance and naivety. It has t be considered also as the driving force in the evolution of technology, which is the crystallization of a way of thinking about the functions of media and about the nature of being human.

In the worst case, the ideal cycle of communication can be used to mimic the infamous hypodermic needle model, i.e., using the outputs from the monitoring AI system to create AI dynamic real-time synthetic content to reach a specific goal in the variables that are measured by the monitoring system. It means that the subject can be 'controlled' by the two AI systems, reception and production, and that the emotional, cognitive outputs can be set to certain thresholds through the exposure to contents that have been generated in real-time to achieve this task. On the contrary, the most positives of the uses of this cycle can be focused in the treatment of certain diseases or cognitive disorders, with AI dynamically creating contents that are dependent on the individuals' responses to the stimuli. In both cases, we accept the hypotheses that the exposure to stimuli (media contents) is capable of having an impact on the variables we are measuring, and these variables are connected to well-known aspects of the human mind and cognition.

## The parasocial interaction AI-mediated effect. An exemplification of how the IMAGINE model works.

As it has been already noted, the embedment of AI in the whole process of communication is expected to have a huge impact on the methods and measures of the effects of media exposure on people. One of the more established media effects theories is that of parasocial interaction (PSI), the perception of an individual to have a friendship relationship with a character in media, usually a celebrity (Horton & Wohl, 1956; Giles, 2002). From the 1970s onwards, several scales have been proposed to measure people's parasocial interactions and relationships with characters in media (Levy, 1979; Rubin, Perse & Powell, 1985; Auter & Palmgreen, 2000; Cohen, 2001; Schramm & Hartmann, 2008). Researchers have found different factors that have an impact on the intensity of parasocial interactions, such as attractiveness, similarity, selective exposure, gender, age or education (Klimmt, Hartmann & Schramm, 2006). Although survey and the use of scales to measure the intensity of the interaction has been the most widely used method in parasocial interaction research, in the last decades there has been a boost in experimentation. Researchers have also used experimentation for measuring the impact of directly addressing viewers in TV series (Cohen, Oliver & Bilandzic, 2019) or in educational videos (Beege, Schneider, Nebel & Rey, 2017), or the influence of PSI on people's likelihood to agree with the source's message (Nah, 2022). However, experimentation has been particularly used in relation to digital technologies, and the role played by virtual reality, avatars, and influencers in social networks. Among them, experimental researches on the impact of interdependent self-construal on parasocial interaction with avatars in the console game Wii (Jin & Park, 2009), the impact of PSI with avatar on cognition and affects (Jin, 2011), the effects of body shape and popularity level on the PSI (Jin, 2018), the impact of 3D sound on social presence and on parasocial interaction (Shin, Song, Kim & Biocca, 2019), on the influence of how streamers address participants and their attention to the chat on experiences of PSI (Wulf, Schneider & Queck, 2021), on the relationship between consumers and AI voice assistants and the evaluation of recommended products (Whang & Im, 2021), the influence of VR headsets on PSI (Kang, Dove, Ebright, Morales & Kim, 2021), the influence of PSI on consumer-AI chatbot interaction (Youn & Jin, 2021), the influence of PSI on successful influencer marketing in Instagram (Kim, 2022), the influence of PSI in the VR shopping environment on brand equity, or the interaction of audiences with virtual influencers compared to real influencers (Stein, Linda Breves & Anders, 2022).

The face is one of the most powerful and dynamic instruments in social communication (Jack & Schyns, 2015), and its attractiveness is considered to be one of the key variables in the process of parasocial interaction and relationship. There are several features that can contribute to face attractiveness, such as symmetry, averageness or sexually dimorphic treats (Fink & Penton-Voak, 2002). However, while some expressions as those of dominance and trustworthiness can be easily manipulated by means of digital dynamic masks, attractiveness remains difficult to camouflage, becoming a source of inequality (Jack & Schyns, 2015). Attractiveness is a characteristic of almost all media celebrities (Gong & Li, 2017), and one of the main requirements for an influencer to be successful (Wiedmann & von Mettenheim, 2021).

The development of algorithms that can automatically and simultaneously model social traits for synthetic faces related to aesthetic, mood and personality (Fuentes-Hurtado, Diego-Mas, Naranjo & Alcañiz, 2018) are trying to overcome this inequality. In particular, is has been shown that avatar's attractiveness can have behavioural consequences, such as effective learning (Wang & Yeh, 2013), the help provided to the avatar by users in online games (Waddell & Ivory, 2015), the increase of the performance in online games (Yee, Bailenson & Ducheneaut, 2009) or in completing a game (Dubosc, Gorisse, Christmann, Fleury, Poinsot & Richir, 2021), the willingness to buy a product (Lee, Hong & Park, 2021), or online gamers' loyalty (Liao, Cheng & Teng, 2019).

Transforming one's virtual aspect has been a trend for years in social networks (Hedman, Skepetzis, Hernandez-Diaz, Bigun & Alonso-Fernandez, 2021). By means of algorithms, users can totally alter their face and body image even in real-time (Shah and Allen, 2019; Thies, Zollhofer, Stamminger, Theobalt & Niessner, 2016). As a general norm, the transformation of one's image follows attractiveness as an objective, since face attractiveness is considered to have a positive influence on one's social success (Little, Jones & DeBruine, 2011). Several digital techniques of efficient face beautification have been developed to increase people's portraits facial attractiveness (Leyvand, Cohen-Or, Dror & Lischinski, 2008; Hu, Shum, Liang, Li & Aslam, 2021), or to totally transform one's face aspect (Perov et al., 2020). This process has been compared to make up or to cosmetic surgery in the physical world (Liu et al., 2019).

Consequently, it is suitable to state that current AI technologies provide the capacity to create and transform faces in digital and virtual spaces in real-time with the objective of optimizing the process of parasocial interaction. This optimization can lead to the increase of the perceived credibility and trustworthiness of the source-face, and to the promotion of certain behavioural outcomes, such as buying or voting. At the same time, the cognitive and emotional influence of the source-face can be measured by AI technologies of reception.

So, the conditions under which interaction takes place follow the characteristics of the IMAGINE model. Both the process of creation of the face and the measure of the responses to it are performed by algorithms that use several sources of data to perform their operations. The process is integrated and bidirectional, in the sense that the outputs measured by the AA-receptor, linked to likeableness, credibility or trustworthiness, are used as an input for the AA-negotiator, that is the responsible for creating the inputs for the AA-creator. The process is performed in real-time, so any change in the reception has an impact on the creation. The whole exposure to the source has a goal that can be determined by the AA-receptor; the whole process is defined to optimize the results of the exposure in relation to the goals in the AA-negotiator. The process is adaptable, and the changes in the different variables in play in the process of creation is continuous, with the aim to fine-tune reception. The characteristics of dynamism and adaptation turn the message the individual is exposed to into something ephemeral and unique, that can't be repeated. Finally, the whole process is assumed to be non-invasive, by means of the technologies for assessing the impact of the exposure to media contents already mentioned.

## Mirroring face-to-face communication

The use of AI technologies to create human-like characters that are treated as friends (Donath, 2021) can be interpreted as a final step to reproduce face-to-face interactions. As already observed by Rubin, Perse and Powell (1985), characters in media encourage parasocial relationships by mirror interpersonal communication. The popularisation of the internet has made possible the increase of social interactions online (Parks & Floyd, 1996). So, we have been observing similar processes of interaction with celebrities and influencers, and finally with robots or virtual influencers, with which people interact not as with an imagined programmer, but with a real social actor (Sundar & Nass, 2000). Consequently, both the design of AI entities or the transformation of oneself online will be shaped by the impact of attractiveness on success (Rand & Hall, 1983), income (Cunningham, 1986), increase of social support or professional access (Bradshaw & DelPriore, 2022), credibility, likeability (Endacott & Leonardi, 2022) or people's emotions (Rodero, Larrea & Mas, 2022). Social cues, such as facial expressions or human-like voices make artificial intelligence actors to look more human (Dehnert & Mongeau, 2022).

## Limitations

Crawford criticises the lack of criticism among researchers to Ekman's findings, considering the scientific controversy (Barret, Marsella, Martinez & Pollak, 2019). Corporations and institutions interested in investing in facial expressions and emotions, although scientific findings show that most of the conclusions derived from Ekman's works and those studies in computer science that have followed his results are, in general, wrong or not accurate enough to conclude that emotions can be easily inferred from people's facial expressions. As Crawford (2021) concludes, in spite of the flaws and failures of the methodology, corporations continue in their seek of sources for mining facial expressions.

Barrett et al. (2019) point out that it is a mistake to conclude what people feel from facial emotions, and claim that context must be considered. However, this appellation to context may seem a common view too for computer scientist in computer vision, since different complementary techniques are being, or will be developed in the future that can overcome the problems of facial expressions recognition, that will be able to capture the nuances of people's most internal feelings and emotions.

Ekman (2016) is well known for his studies on facial expressions that, according to Crawford (2021) are connected to intelligence fundings and US security programs, a pattern that can be extended to other countries, in particular China, Russia or Israel, in what the author describes as a combination of ideology, fear-based politics, and what is more important, the desire to collect information about individuals, without their permission.

The capture of millions of faces from social networks by technology companies, the promise of uncovering the interior emotions or personality traits by means of machine earning technologies.

Executives, who usually ignore statistics and data, are those who exaggerate the power of data and methods, and those executives, as Hammerbacher (2016, cited in Elish and boyd) asserts, repeat the same lies, until they are true. The question is not about the classical propaganda techniques; it is about constructing an idea, a promise, of what data should do, or what we should do with data. And it activates, promotes, reinforces the strategic driver of technology development.

Elish and boyd (2018) assert that discourses on the power of AI is an expression of long-term goals, promises for attracting investments, it justifies the development of methods that are in tune with a certain culture or ideology, defining what is culturally and socially desirable, so that the market positively responds to it, but hide what is behind.

The author is no stranger to the overpromising nature of AI technologies (Elish & boyd, 2018), or to the ontological and epistemological questions when it comes causality and to the measure of complex human emotions or beliefs, and the causality when dealing difficulties to measure complex human emotions or beliefs (Baran & Davis, 2009).

## Conclusion

In his famous Understanding media, Marshall McLuhan (1987) followed Bertrand Russell's and A. N. Whitehead's thinking on the discoveries of the last two centuries to conclude that what is important in arts nowadays is the effect, and afterwards it comes the poem, the painting or the media content that is pretended to provoke the mentioned effect. AI is a step forward in this process of anticipation of the effect, and it has the power to radically change the nature of mediation. Media researchers may be ready for the challenge it will represent.

The IMAGINE model proposed has been guiding line in the evolution of media technology in history. And its crystallization is only possible in a context in which AI controls all the stages of the communication process. IMAGINE allows the conceptualization of a communicative situation in which media contents are AI dynamically created and distributed in real-time (AA-creator) to achieve certain goals that are measured and controlled synchronically, in real-time, by AI technologies of reception (AA-receptor).

The model is, obviously, based on the principles of complete understanding of human responses to external stimuli, and the capacity of AI technologies to create narratives as if they were human. However, we are still far from this scenario, and many scholars claim that the power of AI has been exaggerated, and that certain technologies, such as those of face emotion recognition, are far from providing acceptable results for scientists (Cabitza, Campagner, Mattioli, 2022).

The creation or modification of contents in real time using AI, and as a response of the values of the variables that can be constantly tracked by AI, may allow the media industry to adjust or modulate these variables in a continuous process. Thus, experiments with discrete variables that can have a limited number of possible values may be replaced by a continuous transformation of the content with the objective of modulating the measured variables that are related to the pretended outcomes of individual's exposure to media. This idea is also linked to the aforementioned conception of social physics, but also to that of medicine. The individual is understood as a patient to whom different drugs are administered to control certain variables that are key for the stabilization of her body.

Add

expression shown while delivering the speech determines the effectiveness of the communication and can be very influential in organizational settings

The study included two brown-haired/dark-eyed male and female and two blonde-haired/blue-eyed male and female presenters to investigate the effect of appearance and gender.

## Acknowledgements